\pdfoutput=1
\PassOptionsToPackage{dvipsnames, svgnames, x11names}{xcolor} 
\documentclass[11pt]{article}

\usepackage[preprint]{acl}

\usepackage{times}
\usepackage{latexsym}
\usepackage[T1]{fontenc}
\usepackage[utf8]{inputenc}

\usepackage{microtype}

\usepackage{multirow}
\usepackage{multicol}
\usepackage{geometry}
\usepackage{inconsolata}
\usepackage{caption}
\usepackage{amsfonts}
\usepackage{booktabs}
\usepackage{xspace}
\usepackage{comment}
\usepackage{framed}
\usepackage{enumitem}
\usepackage{url}

\usepackage{graphicx}
\usepackage{tcolorbox}
\usepackage{amsmath}
\usepackage{subfigure}
\newcommand\newfootnote[1]{%
  \begingroup
  \renewcommand\thefootnote{}\footnote{#1}%
  \addtocounter{footnote}{-1}%
  \endgroup
}

\title{MSMO-ABSA: Multi-Scale and Multi-Objective Optimization for Cross-Lingual Aspect-Based Sentiment Analysis}

\author{
  \textbf{Chengyan Wu$^\ast$\textsuperscript{1}},~
    \textbf{Bolei Ma$^\ast$\textsuperscript{2}},~
  \textbf{Ningyuan Deng\textsuperscript{3}},~
  \textbf{Yanqing He$^\ddagger$\textsuperscript{3}},~
  \textbf{Yun Xue\textsuperscript{1}, 
  \textbf{Xiaoyong Liu$^\ddagger$\textsuperscript{4,5}}}
\vspace{3.5pt}
\\
\textsuperscript{1}South China Normal University, Foshan, China~~~\\
\textsuperscript{2}LMU Munich \& Munich Center for Machine Learning, Munich, Germany~~~\\
\textsuperscript{3}Institute of	Scientific and Technical Information of China, Beijing, China~~~\\
\textsuperscript{4}Guangdong Polytechnic Normal University, Heyuan, China~~~\\
\textsuperscript{5}Jilin Engineering Normal University, Changchun, China~~~
\vspace{3.5pt}
\\
\small{
   \texttt{\{chengyan.wu,xueyun\}@m.scnu.edu.cn, bolei.ma@lmu.de, \{dengny2022,heyq\}@istic.ac.cn}, liuxy@gpnu.edu.cn
}
}

\begin{document}
\maketitle
\begin{abstract} 
Aspect-based sentiment analysis (ABSA)
garnered growing research interest in multilingual contexts in the past. 
However, the majority of the studies lack more robust feature alignment and finer aspect-level alignment. 
\newfootnote{$^\ast$Equal contributions.}
\newfootnote{$^\ddagger$Corresponding authors.}
In this paper, we propose a novel framework, \textbf{MSMO}: \textbf{M}ulti-\textbf{S}cale and \textbf{M}ulti-\textbf{O}bjective optimization for cross-lingual ABSA. During multi-scale alignment, we achieve cross-lingual sentence-level and aspect-level alignment, aligning features of aspect terms in different contextual environments. Specifically, we introduce code-switched bilingual sentences into the language discriminator and consistency training modules to enhance the model's robustness. During multi-objective optimization, we design two optimization objectives: supervised training and consistency training, aiming to enhance cross-lingual semantic alignment. To further improve model performance, we incorporate distilled knowledge of the target language into the model.
Results show that MSMO significantly enhances cross-lingual ABSA by achieving state-of-the-art performance across multiple languages and models.
\footnote{Code at \url{https://github.com/swaggy66/MSMO}}
\end{abstract}

\section{Introduction}

Aspect-based sentiment analysis (ABSA) involves identifying specific aspect terms and their sentiment polarity within a sentence \citep{liu2012,pontiki2014semeval}. While research in ABSA has seen success with English texts, 
real-world social media interactions often involve multiple languages \citep{mao2022, zhang2021towards}, highlighting the need for cross-lingual sentiment analysis.
\begin{figure}[htbp]
\centering
  \includegraphics[width=0.98\columnwidth]{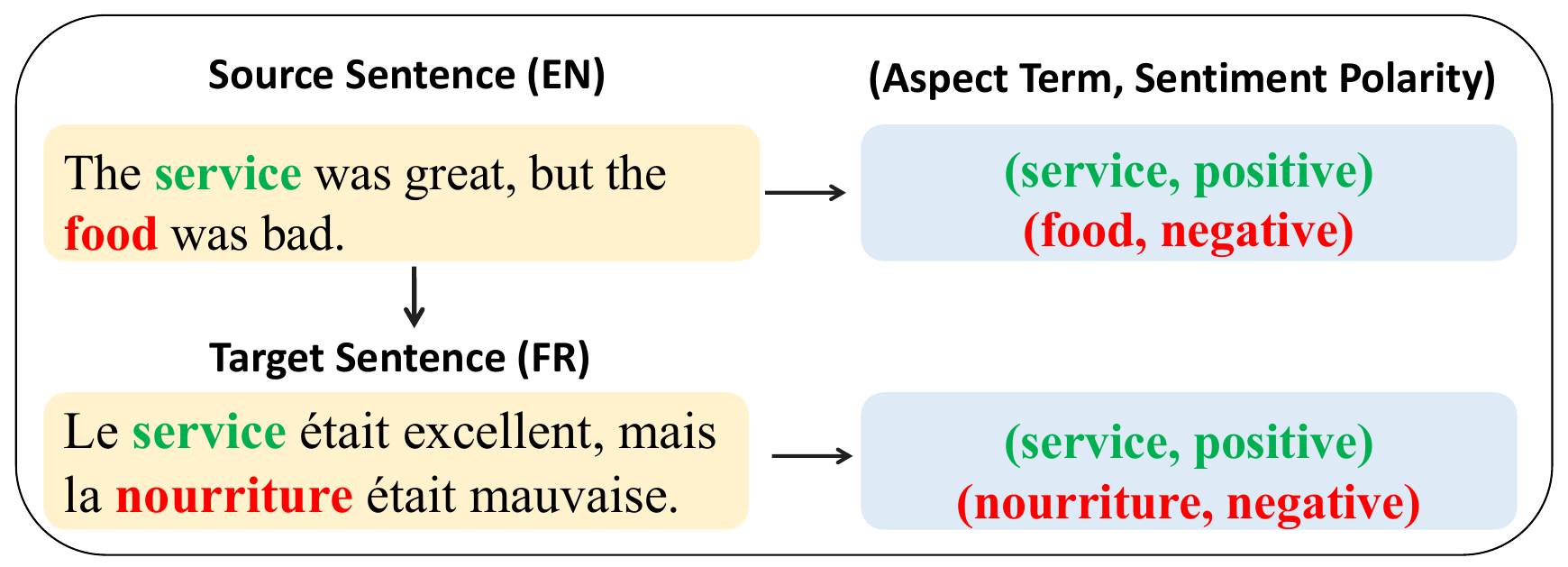}
  \caption{An example of a cross-lingual ABSA task. We train on the source language and perform aspect term extraction and sentiment polarity prediction on the target language.}
  \label{fig:experiments}
\end{figure}
For example, as illustrated in Figure \ref{fig:experiments}, if English is the source language and French is the target, a model trained on a labeled dataset in English should be able to identify the aspect terms ``service'' and ``nourriture'' in the French sentence, with sentiments ``positive'' and ``negative'' respectively.

Since obtaining large amounts of annotated training data for low-resource languages is extremely expensive, early cross-lingual sentiment analysis efforts \citep{zhou2016,xu2017,barnes2018} solely rely on annotated data from different source languages to learn sentiment classification for target languages. These models typically depend on bilingual dictionaries, pre-trained cross-lingual word embeddings, or machine translation to bridge the gap between source and target languages. 

With the advent of multilingual pre-trained language models, recent research has shifted focus to data-level alignment, leveraging multilingual pre-trained models to fine-tune aligned translated data to bridge the gap between source and target languages \cite{li2021unsupervised,zhang2021cross,Bigoulaeva_Hangya_Gurevych_Fraser_2023,lin2023}.

To further enhance the awareness of the multilingual ABSA task, and inspired by previous adversarial training \cite{Wang_Pan_2018,zhou-etal-2022-conner} and consistency training methods \cite{wang2021unsupervised,zhou-etal-2022-conner},
we propose a framework based on \textbf{m}ulti-\textbf{s}cale and \textbf{m}ulti-\textbf{o}bjective optimization, called \textbf{MSMO}. Specifically, the MSMO framework comprises four key components: a feature extractor, a language discriminator, a consistency training module, and a sentiment classifier. 
For the multi-scale aspect, we employ adversarial training for sentence-level alignment and consistency training for aspect-level alignment, leveraging both the bilingual translated dataset and the code-switched dataset in the process. Specifically, introducing a code-switched dataset that switches different aspect terms can introduce perturbations, allowing the embedding spaces of the source and target languages to align better with the anchor aspects and improving the robustness of the model. 
For the multi-objective optimization, we combine supervised training and consistency training as optimization objectives, aiming to align aspect terms of different languages at a finer granularity. Additionally, we extend the MSMO framework to multilingual ABSA. To explore the importance of unlabeled target language knowledge for performance improvement, we also apply knowledge distillation using unlabeled data in the target language.

In summary, our main contributions are:

\begin{itemize}[leftmargin=*,nolistsep]
\item\textbf{Sentence-level Alignment}: We propose an adversarial training approach using a code-switched dataset. This method enhances the language discriminator's ability to capture invariant features and develop more robust representations across languages by introducing aspect term perturbations.
\item\textbf{Aspect-level Alignment}: We utilize consistency training to ensure the model provides consistency predictions for aspect terms with the same sentiment, improving alignment at the aspect level.
\item\textbf{Multi-objective Optimization}: We integrate supervised training and consistency training objectives to minimize the performance gap between different languages.
\item\textbf{Extensive Evaluation}: We conduct comprehensive experiments on benchmark datasets in five languages across cross-lingual and multilingual settings. The results demonstrate that our MSMO achieves state-of-the-art performance.
\end{itemize}

\section{MSMO Framework}

\begin{figure*}[htbp]
  \centering
   \includegraphics[width=\textwidth]{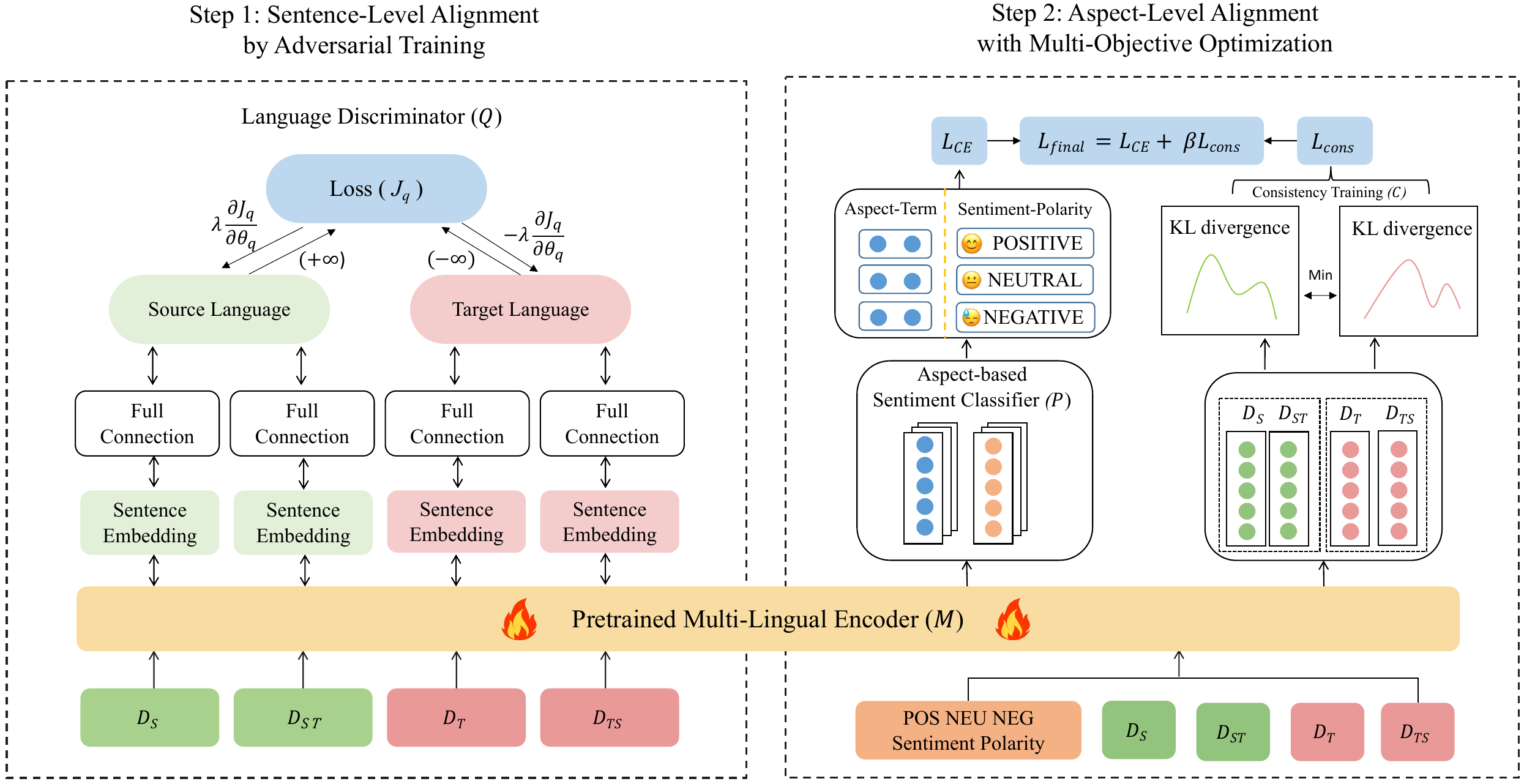}
  \caption{The MSMO framework. It mainly comprises two basic steps:  (1). Sentence-level alignment by adversarial training (\S\ref{sec:step1}); (2).  Aspect-level alignment with multi-objective optimization (\S\ref{sec:step2}). The \colorbox{Moccasin}{Pretrained Multi-Lingual Encoder} connects both steps by updating the parameters from the loss of language discriminator in step 1 and from the combined loss in step 2. }
  \label{fig:msmo}
\end{figure*}

\subsection{Problem Formulation and Background}
We regard the ABSA task as a sequence labeling problem. Given an input sentence 
\( x = \{x_1, x_2, \ldots, x_n\} \) containing \( n \) tokens, our goal is to 
predict the labels \( y = \{y_i\}_{i=1}^n \) for the input sequence, where 
\( y_i \in Y = \{B, I, E, S\}\text{-}\{POS, NEU, NEG\} \cup \{O\} \), 
representing the aspect term boundaries and their sentiment polarities corresponding to the token \( x_i \). 

In our cross-lingual transfer framework, following \citet{zhang2021cross}, we use the source language data \( D_{S} \) and the translated target language data \( D_{T} \).
We also use the code-switched data 
(\( D_{S_{T}}, D_{T_{S}} \)) during training, where $D_{S_{T}}$ is created by replacing the aspect terms in $D_{S}$ with their counterparts in the target language, and $D_{T_{S}}$ is created by replacing the aspect terms in $D_{T}$ with their counterparts in the source language. The training data \( D_U \) consists of sentence-label pairs 
\( (x_u, y_u) \in D_U \), where $D_S\cup D_T\cup D_{S_{T}}\cup D_{T_{S}} \in D_U$, aiming to predict the label sequence \( y_{t} \) for the target language in the test set. 

\subsection{Preliminaries of the MSMO Framework}
\label{sec:preliminaries}

Figure \ref{fig:msmo} illustrates the key components of our method, which mainly consists of two overall steps: (1). Sentence-level alignment by adversarial training (\S\ref{sec:step1}); (2).  Aspect-level alignment with multi-objective optimization (\S\ref{sec:step2}). Before delving into the details of the two steps, we illustrate the preliminaries of the \colorbox{Moccasin}{Pretrained Multi-Lingual Encoder}.

The MSMO framework integrates a pre-trained multilingual encoder (\( M \)) as a feature extractor to generate contextual representations of sentence tokens. 
Given a sequence of \( n \) tokens \([x_1, x_2, \ldots, x_n]\), we take the final hidden layer outputs of the \( M \) as the intermediate representations \( h_i \in \mathbb{R}^{l} \). 
\begin{equation}
[h_1, h_2, \ldots, h_n] = \text{\( M \)}([x_1, x_2, \ldots, x_n])
\end{equation}

These intermediate token embeddings $h_i$ are then fed into two different branches. In the first stage (\S \ref{sec:step1}), $h_i$ is input into the language discriminator $Q$, which aims to predict a scalar score indicating whether $x$ is from the source or the target. In the second stage (\S \ref{sec:step2}), we use the updated \( M \) from the first stage to encode and input the representations into the sentiment classifier $P$ and the consistency training module $C$. $Q$ uses a sigmoid activation function, while both $P$ and $C$ use softmax activation functions, defined as:
\begin{equation}
p(x_i) = F(\text{Dropout}(W*h_i + b))
\end{equation}

$P$ predicts labels for the input sequences based on the feature representation $h_i$, while $C$ aligns aspect terms with the same sentiment polarity across different languages by aligning the predicted probability distributions of aspect terms in the source language and target language.

\subsection{Step 1: Sentence-Level Alignment by Adversarial Training}
\label{sec:step1}

Now we look into the first step which leverages a language discriminator to conduct sentence-level alignment by adversarial training. In previous research, ADAN \citep{chen-etal-2018-adversarial} is proposed to use the Wasserstein distance \citep{arjovsky2017} for standard adversarial training, addressing the instability of adversarial training in the ADAN-GRL \citep{ganin2015domain} method and achieving better performance. Inspired by the ADAN paradigm, we design a language discriminator \( Q \) to reduce the semantic gap across languages. \( Q \) is a binary classifier with a sigmoid layer on top, so the language recognition score is always between 0 and 1, aiming to determine the probability that the input text \( x \) is from source or target based on the hidden features \( h_i \) captured by the feature extractor.
For training, \( Q \) is connected to the encoder via a gradient reversal layer \citep{ganin2015}, which retains the input during the forward pass but multiplies the gradient by \(-\lambda\) during the backward pass (we set $\lambda$ to 1 in Figure \ref{fig:msmo}). 
Thus, standard backpropagation can be used to train the entire network holistically. However, unlike ADAN, to enhance the robustness of the model while preserving semantic consistency, we also introduce the code-switched dataset in addition to using bilingual parallel corpora. Specifically, for source language sentences that introduce aspect terms in the target language, we stimulate the language discriminator \( Q \) to identify them as source language sentences, and \emph{vice versa}. 

By inducing local perturbations from aspect term changes in sentences, we encourage the language discriminator to recognize common features of languages and stimulate the feature extractor to capture invariant features of languages through backpropagation.  Our objective is to approximately minimize the Wasserstein distance between $(P(h_i), P(h_i'))$ according to the Kantorovich-Rubinstein duality \citep{Villani_2013}. To ensure that \( Q \) is a Lipschitz function (up to a constant), the parameters of \( Q \) are always clipped to a fixed range \([-c, c]\). Thus, the objective $J_q$ of $Q$ becomes:

\noindent
{\footnotesize
\begin{multline}
\label{eq:j}
J_q(P(h_i), P(h_i')) \equiv  \max_{\theta_q} \mathbb{E}[\mathcal{Q}(\mathcal{P}(h_i))] 
- \mathbb{E}[\mathcal{Q}(\mathcal{P}(h_i'))]
\end{multline}}

where $h_i \in D_S \cup D_{S_{T}}$, and $h_i' \in D_T \cup D_{T_{S}}$. The supremum (maximum) of $J_q$ is taken over the set of all 1-Lipschitz functions $Q$. Intuitively, $Q$ tries to output higher scores for source instances and lower scores for target instances. More formally, $J_q$ is an approximation of the Wasserstein distance between $P(h_i)$ and $P(h_i')$ in Equation \ref{eq:j}.

\subsection{Step 2: Aspect-Level Alignment with Multi-Objective Optimization}

\label{sec:step2}

After the first stage of training, the updated encoder has learned to extract invariant features across different languages. We use the updated encoder for the second stage of training, where the extracted features are fed into the sentiment classifier $P$ and the consistency training module $C$.

\paragraph{Supervised Training.}

For the sentiment classifier $P$, we use the traditional cross-entropy loss, denoted as $\mathcal{L}_{CE}$, which is computed between the predicted label distribution and the gold label in one-hot encoding. Therefore, we seek to minimize the following loss function for $P$:

{\small
\begin{equation}
\mathcal{L}_{CE} = \frac{1}{|D_U|} \sum_{(\boldsymbol{x}, \boldsymbol{y}) \in D_U} \left[-\frac{1}{L} \sum_{i=1}^{L} y_{i} \log p_{\theta}\left(y_{i} \mid x_{i}\right)\right]
\end{equation}}
 
where \( L \) represents the length of the sentence \( X \), \( i \) denotes the \( i \)-th token in the sentence. In addition, $(x, y)$ belongs to the labeled training dataset $D_U$.

\paragraph{Consistency Training.}
Consistency training \citep{miyato2019,clark2018,xie2020} aims to reduce the model's overfitting and bias towards specific input forms by guiding the model to produce consistency predictions under different input perturbations. Although it has been successful in CV \citep{wang2024learning} and sentence-level NLP tasks \citep{miyato2019,xie2020}, there has been a lack of effective attempts at consistency training for the cross-lingual ABSA task. In cross-lingual ABSA, learning aspect-level feature representations across different languages can achieve cross-lingual adaptability from a finer-grained perspective, where the model should maintain consistency in the predictions of aspect terms with the same sentiment polarity. 

We explore a consistency training method for cross-lingual ABSA to improve alignment at the aspect level, as follows. Let $\phi$ be a transformation function that generates small perturbations, such as noise from text translation or aspect term transformation. In this paper, one transformation method is to translate $X$ into another language, and another is to swap the aspect terms in the source and target language sequences. Given a sequence of tokens $X$ and aspect terms in the sequence denoted as $s$, we apply the $\phi$ transformation to the source language sequence to obtain a perturbed sequence $X'$, where the aspect terms $s$ after transformation correspond to $s'$. We encourage the model to capture feature representations of aspect terms with the same sentiment across different languages and use bidirectional KL divergence to compute the divergence $D_{\text{div}}$ between the probability distributions of the aspect term pairs $(s, s')$ at the span level, and then minimize their consistency loss $\mathcal{L}_{\text{cons}}$ to output consistent probability distributions over $s$ and $s'$:

\noindent
{\small
\begin{multline}
\mathcal{L}_{\text{cons}} = \frac{1}{m} \sum_{{(s_i,s_i')} \in \mathbf{(X,X')}} \frac{1}{2} \big[ \operatorname{KL} \left(P \left(y_i' \mid s_i' \right) \,\middle\|\, P \left(y_i \mid s_i \right) \right) \\ 
+ \operatorname{KL} \left(P \left(y_i \mid s_i \right) \,\middle\|\, P \left(y_i' \mid s_i' \right) \right) \big]
\end{multline}
}

where $y_i$ and $y_i'$ are the labels of the spans and $m$ is the total number of aspect term pairs $(s_i, s_i')$. Inspired by the approach of \citet{zhou-etal-2022-conner}, we define the probability of a span as the product of the tokens that constitute the span.

\paragraph{Multi-Objective Optimization.}
As described above, we apply supervised training and consistency training to the source language data, translated target language data, and code-switched data. Then, we combine the cross-entropy loss $\mathcal{L}_{\text{CE}}$ with the consistency loss $\mathcal{L}_{\text{cons}}$ to form our total training objective:

\begin{equation}
\label{eq:l}
\mathcal{L}_{\text{total}} = \sum_{X \in D_{\text{U}}} \mathcal{L}_{\text{CE}}   + \sum_{X \in D_{\text{U}}}  \beta \mathcal{L}_{\text{cons}}
\end{equation}

\section{Experimental Setups}
\subsection{Dataset}
We choose the SemEval-2016 dataset \citep{pontiki2016semeval} to evaluate our method. 
This dataset consists of real user reviews across eight languages, with ABSA annotations available for English (EN), French (FR), Spanish (ES), Dutch (NL), Russian (RU), and Turkish (TK). However, due to the limited size of the Turkish test set (fewer than 150 sentences), we excluded it from our evaluation, consistent with prior multilingual ABSA studies \cite{zhang2021cross, lin2023}. For a fair comparison, we use the data processed by \citet{zhang2021cross}. The data for each language is divided into training, validation, and test sets, along with a code-switched dataset. We use English as the source language during training, and the other languages as target languages in the prediction phase.

We utilize the code-switched dataset proposed by \citet{zhang2021cross}, referred to as $D_{S_{T}}$ and $D_{T_{S}}$. $D_{S_{T}}$ is constructed by replacing the aspect terms in the source language dataset $D_S$ with aspect terms that appear in the target language dataset $D_T$. In contrast, $D_{T_{S}}$ is generated by replacing the aspect terms in the target language dataset $D_T$ with aspect terms that appear in the source dataset $D_S$. 

\begin{figure}[htbp]
\centering
  \includegraphics[width=0.98\columnwidth]{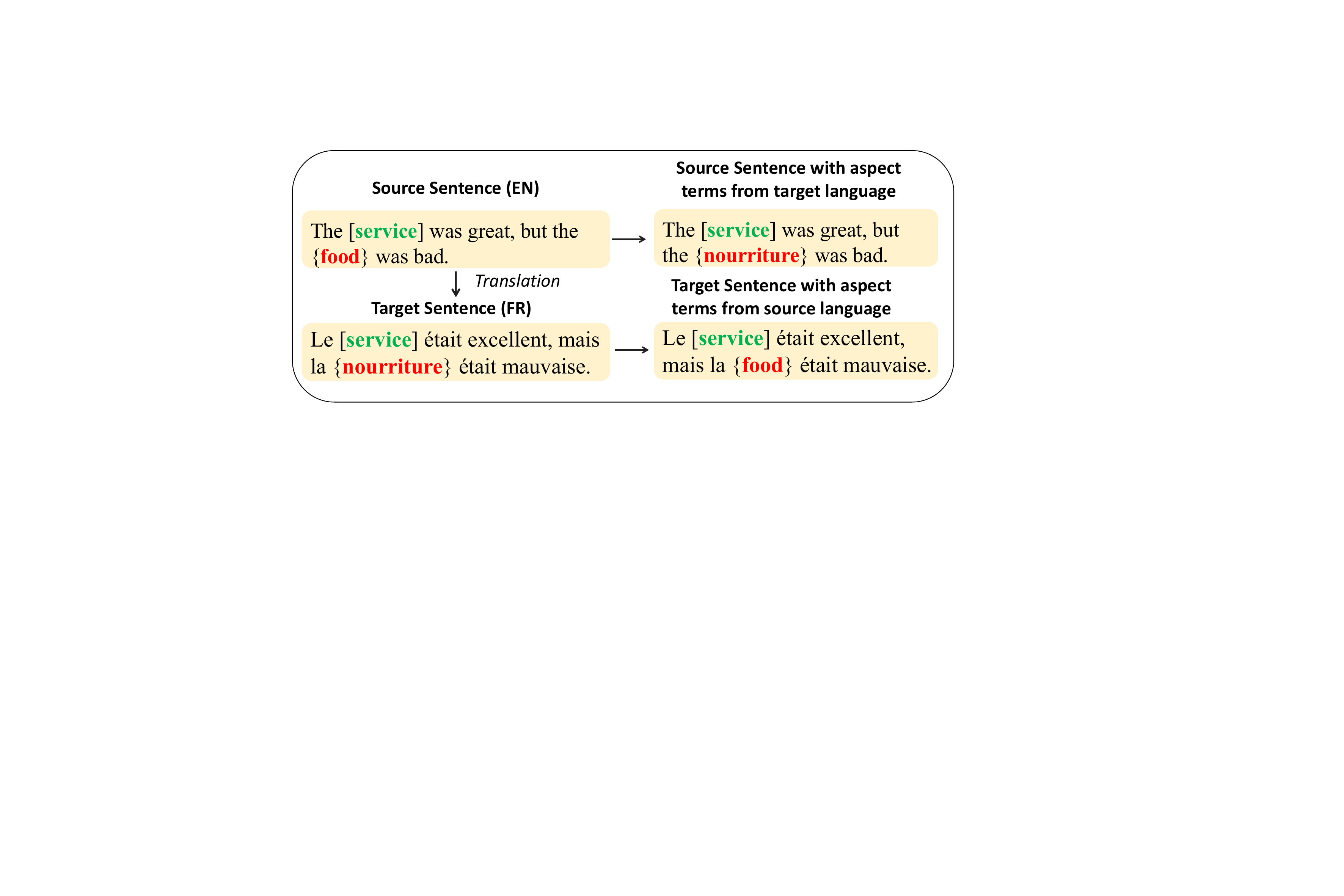}
  \caption{An example of the code-switched dataset.}
  \label{fig:code-switch}
\end{figure}

Figure \ref{fig:code-switch} shows examples of the source language, translated, and code-switched datasets. The aspect terms ``service'' and ``food'' in the source language sentence \(D_S\) are marked with special symbols (e.g., ``\{\}'', ``[]''). The corresponding target language sentence \(D_T\) is obtained through machine translation, with the aspect terms being ``service'' and ``la nourriture''. Then, by matching the special symbol markers, we can swap the aspect terms between the source and target language sentences. That is, ``service'' and ``food'' in the source language sentence is replaced with ``service'' and ``la nourriture'' in the target language to obtain \(D_{S_{T}}\), and vice versa for the target language sentence to obtain \(D_{T_{s}}\), thereby generating the code-switched data.

The details of the dataset statistics in each language are shown in Table \ref{tab:dataset-semeval}.
\# S and \# A denote the number of sentences and aspect terms in different sets, respectively.

\begin{table}[htbp]
\centering
\footnotesize
\renewcommand\arraystretch{1}
\setlength\tabcolsep{4pt}
\begin{tabular}{lcccccc}
\toprule
          &  &EN&FR&ES&NL&RU    \\ \midrule
\multirow{2}{*}{\textbf{Train} }           & \# S          & 2000 &1664 &2070 &1722 &3655           \\ 
         & \# A     &1743 &1641 &1856 &1231 &3077       \\ \midrule
\multirow{2}{*}{\textbf{Test}}            &\# S     &676 &668 &881 &575& 1209       \\ 
          &\# A      &612 &650 &713 &373 &949       \\ 
\bottomrule
\end{tabular}
\caption{Statistic of the original dataset.}
\label{tab:dataset-semeval}
\end{table}

\subsection{Models and Parameter Settings}
We evaluate four target languages using the Micro-F1 metric on two multilingual pre-trained models, including the cased mBERT \cite{devlin-etal-2019-bert} and the base XLM-R model \cite{conneau-etal-2020-unsupervised}, for a fair comparison with existing methods. A prediction is considered correct only when the tuple (entity, label) is correctly predicted, where the entity is the boundary of the aspect term and the label is the corresponding sentiment polarity. Following the settings of \citet{zhang2021cross}, we set the maximum training steps to 2000 for mBERT and 2500 for XLM-R. Additionally, we allocate different weights for the multi-objective optimization functions of the two models across the four target languages, as in Equation \ref{eq:l}, with $\beta$ values \{4.5e-4, 2.5e-4, 2.5e-4, 3.5e-4\} and \{2.5e-3, 1.5e-3, 1.5e-3, 3.5e-3\} respectively (refer to Appendix \ref{sec:setup_parameter}). Based on the performance of the validation set of the source language, we select the best model in the last 500 steps.

We select the optimal training hyperparameters through a grid search over combinations of batch size and learning rate. The ranges are: learning rate \{1e-5, 2e-5, 5e-5\}; batch size \{8, 16, 25\}. For mBERT, we use a learning rate of 5e-5 and a batch size of 16; for XLM-R, we use a learning rate of 2e-5 and a batch size of 8. For all experiments, we report the average F1 scores over 5 runs with different random seeds.

The MSMO method integrates several advanced components, such as multi-teacher distillation and consistency training, which introduce additional computational demands during both training and inference.  
To quantify this overhead, we report the resources used in our experiments. All experiments were run on a single NVIDIA A6000 GPU (48 GB memory), a CPU with 120 MB (\(\approx 0.12\) GB) of L3 cache, and 42 GB of RAM.

\begin{table}[htbp]
\centering
\footnotesize
\begin{tabular}{l l c}
\toprule
\textbf{Model} & \textbf{Variant} & \textbf{GPU Memory} \\
\midrule
\multirow{3}{*}{mBERT} 
  & MSMO & $\approx$22 \\
  & w/o Language Discriminator & $\approx$18 GB\\
  & w/o Consistency Training   & $\approx$16 GB \\
\midrule
\multirow{3}{*}{XLM-R}
  & MSMO & $\approx$27 \\
  & w/o Language Discriminator & $\approx$24 GB\\
  & w/o Consistency Training   & $\approx$21 GB\\
\bottomrule
\end{tabular}
\caption{Approximate GPU memory usage.}
\label{tab:comp_overhead}
\end{table}

\subsection{Knowledge Distillation Settings}
\label{sec:kd-settings}
\begin{figure}[htbp]
\centering
  \includegraphics[width=0.88\columnwidth]{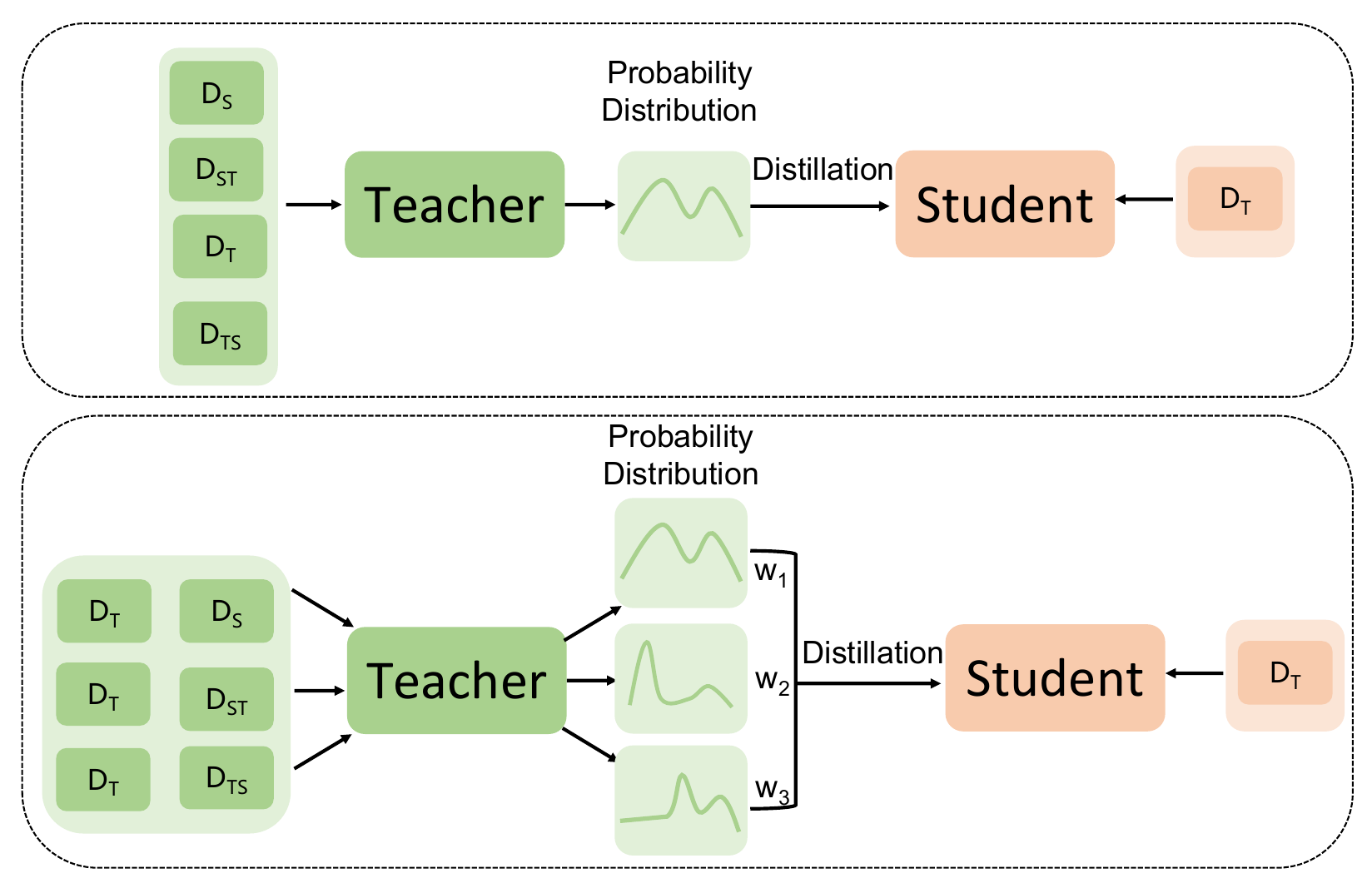}
  \caption{The single-teacher and multi-teacher distillation process.}
  \label{fig:teacher}
\end{figure}

\begin{figure}[htbp]
\centering
  \includegraphics[width=0.88\columnwidth]{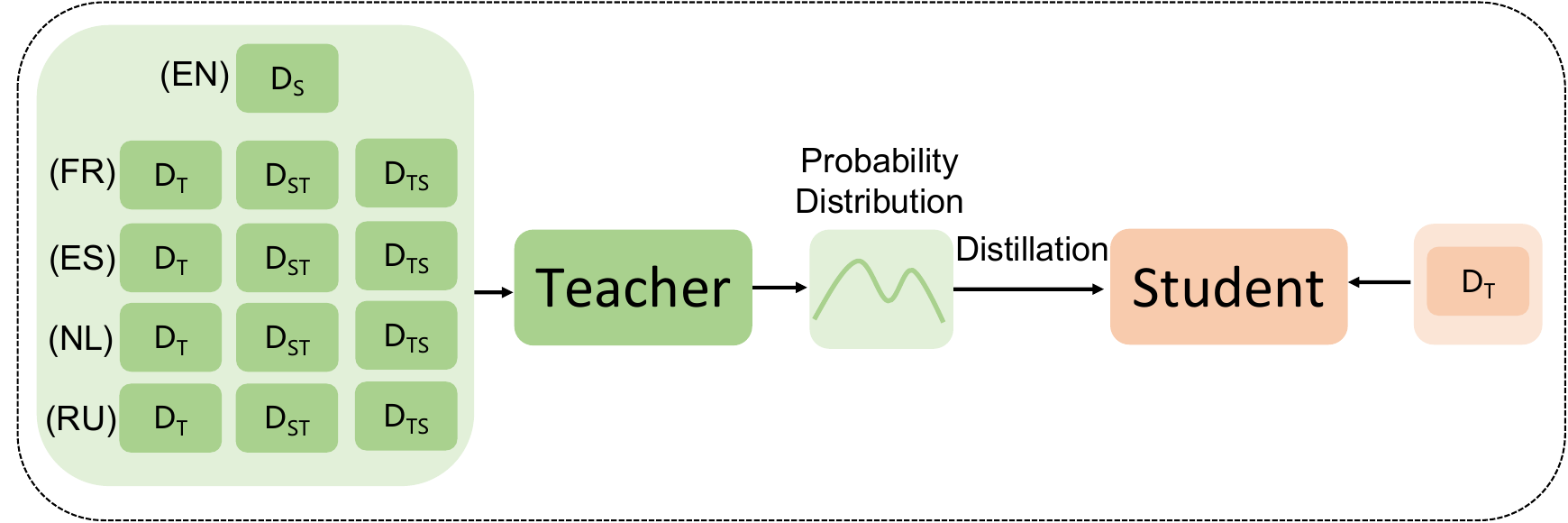}
  \caption{The multilingual distillation process.}
  \label{fig:multilingual}
\end{figure}

Figure \ref{fig:teacher} and \ref{fig:multilingual} illustrate the three modes of knowledge distillation: single-teacher distillation, multi-teacher distillation, and multilingual distillation, proposed by \citet{zhang2021cross}.

For the teacher model, we use a combination of source and target language data (\(D_T\), \(D_S\), \(D_{S_{T}}\), and \(D_{T_{S}}\)) to train the teacher model within our MSMO framework. For the student model, to enable it to perform the ABSA task in the target language, we use the translated target language data as the initial training data, then obtain soft labels for the target language test set predictions from the teacher model, and finally conduct incremental training on this soft-labeled data. For multi-teacher distillation, we assign equal weights to different teacher models, i.e., $w_k = 1/3$ in Equation \ref{eq:weight}. 

{\small
\begin{equation}
\label{eq:weight}
p_{t} = \sum_{k=1}^{3} \omega_{k} * g_{t_{k}}
\end{equation}}
where $w_k$ is the weight for each teacher model. With the combined soft label $g_t$, a student model can be trained similarly by using only the encoder and sentiment classifier modules as in Equation \ref{eq:kd}.

{\small
\begin{multline}
\label{eq:kd}
\mathcal{L}_{KD} = \frac{1}{\left|D_{NL}\right|} \sum_{X \in D_{NL}} \left[ \frac{1}{L} \sum_{i=1}^{L} \operatorname{MSE}\left(p_{t_{i}}, p_{s_{i}}\right) \right]
\end{multline}}

where \( L \) represents the length of the sentence \( X \), and \( i \) denotes the \( i \)-th token in the sentence. Additionally, \( D_{NL} \) indicates the unlabeled dataset in the target language, and \( p_{t_{i}} \) and \( p_{s_{i}} \) are the prediction probabilities of the \( i \)-th token from the student and teacher models, respectively. We use the mean squared error loss \( \mathrm{MSE}(\cdot) \) to measure the difference between the two probability distributions.

\subsection{Compared Methods}

We compare our method against several baseline approaches:

\vspace{5pt}

\begin{itemize}[nolistsep]
    \item \textbf{SUPERVISED}: A fully supervised method where the model is trained using data from the target language.
    \item \textbf{ZERO-SHOT} \citep{conneau-etal-2020-unsupervised}: This method fine-tunes the model on labeled source data and applies it directly to target data, showing strong cross-lingual adaptation.
    \item \textbf{TRANSLATION-TA} and \textbf{BILINGUAL-TA} \citep{li2021unsupervised}: Translation-based. TRANSLATION-TA trains the model with pseudo-labeled data and aligned translations, while BILINGUAL-TA combines source data with aligned translations for training.
    \item \textbf{ACS} \citep{zhang2021cross}: This method introduces code-switched data for training, aligning aspect terms at the data level to bridge the gap between different languages.
    \item \textbf{CL-XABSA} \citep{lin2023}: 
    It performs contrastive learning at both the sentiment levels (SL) and token levels (TL), respectively, minimizing the distance between tokens with the same sentiment and identical labels.
    \item \textbf{Equi-XABSA} \citep{LIN2024125059}: It mitigates sample imbalance and languages’ representation disparity.
\vspace{5pt}
\end{itemize}

Additionally, we evaluate the performance of our method with distilled data compared to:
\vspace{5pt}
\begin{itemize}[nolistsep]
    \item \textbf{Single-Teacher Distillation}: ACS-DISTILL-S and CL-XABSA-DISTILL-S.
    \item \textbf{Multi-Teacher Distillation}: ACS-DISTILL-M and CL-XABSA-DISTILL-M.
\end{itemize}
\vspace{5pt}

\begin{table*}[htbp]
\centering
\scriptsize
\setlength\tabcolsep{7pt}
\begin{tabular}{l|ccccc|ccccc}
\toprule 
\multirow{2}{*}{Methods} & \multicolumn{5}{|c|}{mBERT} & \multicolumn{5}{c}{XLM-R} \\
\cline { 2 - 11 } 
& FR & ES & NL & RU & Avg & FR & ES & NL & RU & Avg \\
\midrule 
SUPERVISED & 61.80 & 67.88 & 56.80 & 58.87 & 61.34 & 67.44 & 71.93 & 64.28 & 64.93 & 67.15 \\
\midrule 
ZERO-SHOT \citep{conneau-etal-2020-unsupervised} & 45.60 & 57.32 & 42.68 & 36.01 & 45.40 & 56.43 & 67.10 & 59.03 & 56.80 & 59.84 \\
TRANSLATION-TA \citep{li2021unsupervised} & 40.76 & 50.74 & 47.13 & 41.67 & 45.08 & 47.00 & 58.10 & 56.19 & 50.34 & 52.91 \\
BILINGUAL-TA \citep{li2021unsupervised} & 41.00 & 51.23 & 49.72 & 43.67 & 46.41 & 49.34 & 61.87 & 58.64 & 52.89 & 55.69 \\
ACS \citep{zhang2021cross} & 49.65 & 59.99 & 51.19 & 52.09 & 53.23 & 59.39 & 67.32 & 62.83 & 60.81 & 62.59 \\
CL-XABSA (SL) \citep{lin2023} &49.75&60.12&49.34&50.10& 52.32&58.10&64.85&59.75&58.84& 60.39\\
CL-XABSA (TL) \citep{lin2023} & 50.55& 60.09&52.45&50.73& 53.46& 59.47& 64.63&59.40&61.13&61.16\\
Equi-XABSA \citep{LIN2024125059} &50.08&63.08&51.85&52.59& 54.40&  60.68&69.56&61.31&62.34& 63.47\\
\textbf{MSMO} & \textbf{51.42} & \textbf{63.26} & \textbf{52.68} & \textbf{53.45} & \textbf{55.20} & \textbf{61.01} & \textbf{69.74} & \textbf{63.26} & \textbf{62.52} & \textbf{64.13} \\
\midrule 
ACS-DISTILL-S \citep{zhang2021cross} & 52.23 & 62.04 & 52.72 & 53.00 & 55.00 & 61.00 & 68.93 & 62.89 & 60.97 & 63.45 \\
ACS-DISTILL-M \citep{zhang2021cross} & 52.25 & 62.91 & 53.40 & 54.58 & 55.79 & 59.90 & 69.24 & 63.74 & 62.02 & 63.73 \\
CL-XABSA-DISTILL-S \citep{lin2023} & 52.76 & 62.54 & 53.38 & 53.48 & 55.27 & 61.20 & 69.13 & 63.01 & 61.37 & 63.68 \\
CL-XABSA-DISTILL-M \citep{lin2023} & 52.99 & 63.54 & 53.52 & 53.98 & 56.01 & 62.10 & 69.37 & 64.27 & 62.29 & 64.51 \\
\textbf{MSMO-DISTILL-S} & \textbf{53.58} & \textbf{63.80} & \textbf{53.97} & \textbf{54.47} & \textbf{56.46} & \textbf{61.69} & \textbf{70.16} & \textbf{63.58} & \textbf{62.96} & \textbf{64.25} \\
\textbf{MSMO-DISTILL-M} & \textbf{54.39} & \textbf{64.59} & \textbf{54.14} & \textbf{54.89} & \textbf{56.94} & \textbf{63.89} & \textbf{69.93} & \textbf{65.15} & \textbf{63.20} & \textbf{65.54} \\
\bottomrule 
\end{tabular}
\caption{Performance comparison of various methods on different languages using mBERT and XLM-R. S denotes single-teacher distillation and M denotes multi-teacher distillation.
} 
\label{tab:performance}
\end{table*}

\section{Results and Analysis}

\subsection{Cross-lingual ABSA Results}
We compare our method with previous methods in Table \ref{tab:performance}. Overall, we achieve SOTA performance on both mBERT and XLM-R models compared to the zero-shot baselines and the CL-XABSA baselines from \citet{lin2023}, indicating that our approach better facilitates semantic convergence across different languages. Additionally, our results are closer to those of the fully supervised fine-tuning method, highlighting the robustness and effectiveness of our approach in bridging the performance gap with supervised methods.

Content-wise, we observe the following in-depth key phenomena: 

\textbf{XLM-R vs. mBERT.} Methods based on the XLM-R backbone generally outperform those based on the mBERT backbone. The primary reason is that XLM-R has a larger number of parameters and uses a larger multilingual corpus during the pre-training phase, leading to stronger cross-lingual adaptation capabilities. 

\textbf{Performance Comparison.} Our proposed MSMO method not only outperforms the ZERO-SHOT method and translation-based methods (BILINGUAL-TA and TRANSLATION-AF) but also achieves better performance than the CL-XABSA method. This demonstrates the effectiveness of introducing sentence-level adversarial training and aspect-level alignment between different languages. 

\textbf{Language-Specific Improvements.} Our MSMO method achieves performance improvements across all four target languages, with a more noticeable improvement in Spanish. Compared to the CL-XABSA method, the performance of MSMO in Spanish improves by 3.14\% and 4.89\% on mBERT and XLM-R, respectively. The primary reason is that languages from different families have different semantic spaces, and Spanish is closer to English in terms of language family, making their semantic spaces more easily converged. This aligns with the findings of \citet{zhang2021cross}.

\textbf{Distillation Performance.} Following the ACS paradigm, we apply the MSMO method to single-teacher distillation (MSMO-DISTILL-S) and multi-teacher distillation (MSMO-DISTILL-M), and both achieve higher performance. This can be explained by the fact that the teacher model of our MSMO method outperforms the teacher model of the CL-XABSA method. During the knowledge distillation process to the target language, the teacher model of the MSMO method can achieve more accurate soft label predictions, thereby better guiding the student model to learn from these soft labels on unlabeled target language data. Additionally, the multi-teacher distillation method performs better than the single-teacher distillation method, which may be because the multi-teacher model can better combine the strengths of different teachers to guide the student model.

\subsection{Multilingual ABSA Results}
To fairly compare with the previous SOTA method by \citet{lin2023}, we report the results of the MSMO method in a multilingual setting (MTL-MSMO) in Table \ref{tab:multilingual_results}. As shown in Table \ref{tab:multilingual_results}, the MTL-MSMO method outperforms the multilingual CL-XABSA method (MTL-CL-XABSA) for the teacher model, with average Micro-F1 improvements of 1.61\% and 0.77\% on the mBERT and XLM-R models, respectively. This indicates that using the MSMO method for distillation with unlabelled data can achieve higher performance. This improvement can be attributed to our language discriminator and consistency training modules, which better align the semantic spaces in a multilingual setting. Furthermore, due to the superior performance of the teacher model, the student model can learn knowledge from multiple target languages through the soft labels predicted by the MTL-MSMO teacher model and apply it to specific language inference. Our method also shows significant improvement in multilingual knowledge distillation, with average Micro-F1 improvements of 0.97\% and 0.77\% on the mBERT and XLM-R models, respectively, over the previous MTL-CL-XABSA-DISTLL, demonstrating the effectiveness and superiority of our proposed method.

\begin{table}[htbp]
\centering
\scriptsize
\setlength\tabcolsep{3pt}
\renewcommand\arraystretch{0.75}
\begin{tabular}{cccccc}
\toprule
& FR & ES & NL & RU & Avg \\
\midrule
\multicolumn{6}{c}{Based on mBERT:} \\
\midrule
MTL-CL-XABSA & 50.01 & 59.05 & 51.22 & 50.59 & 52.72 \\
MTL-CL-XABSA-DISTLL & 53.03 & 62.19 & 54.25 & 54.63 & 56.03 \\
\textbf{MTL-MSMO} & \textbf{51.33} & \textbf{60.43} & \textbf{53.68} & \textbf{51.89} & \textbf{54.33} \\
\textbf{MTL-MSMO-DISTLL} & \textbf{54.56} & \textbf{62.69} & \textbf{55.56} & \textbf{56.19} & \textbf{57.00} \\
\midrule
\multicolumn{6}{c}{Based on XLM-R:} \\
\midrule
MTL-CL-XABSA & 60.09 & 68.88 & 64.16 & 63.07 & 64.05 \\
MTL-CL-XABSA-DISTILL & 62.37 & 70.58 & 65.98 & 62.79 & 65.43 \\
\textbf{MTL-MSMO} & \textbf{60.93} & \textbf{69.34} & \textbf{64.85} & \textbf{64.17} & \textbf{64.82} \\
\textbf{MTL-MSMO-DISTLL} & \textbf{63.23} & \textbf{70.95} & \textbf{66.24} & \textbf{64.36} & \textbf{66.20} \\
\bottomrule
\end{tabular}
\caption{Multilingual results (MTL) with mBERT and XLM-R as backbone respectively. MTL-CL-XABSA and MTL-CL-XABSA-DISTLL are selected from the best result between TL and SL in \citet{lin2023}.}
\label{tab:multilingual_results}
\end{table}

\subsection{Ablation Study}
To demonstrate the effectiveness of the main components in the MSMO method, we experiment with ablations and present the results in Table \ref{tab:ablation_results}. We design two variants of MSMO for the experiments: 

\noindent
1) \textbf{w/o. Language Discriminator}: We remove the language discriminator, retaining only the feature extractor, sentiment classifier, and consistency training modules to train the model; 

\noindent
2) \textbf{w/o. Consistency Training}: We remove the consistency training module, retaining only the feature extractor, sentiment classifier, and language discriminator to train the model.

As shown in Table \ref{tab:ablation_results}, the results indicate that 

\noindent
1) MSMO experiences a performance drop when any component is removed, demonstrating the importance of all components; 

\noindent
2) Removing the language discriminator results in a decrease of 1.68\% and 1.30\% in the average F1 score, respectively, indicating that training the language discriminator with bilingual data and code-switched bilingual data helps the model learn language-invariant features, contributing to performance improvement. Specifically, in different code-switched contexts, the model can better focus on the boundary changes of aspect terms in different languages, thereby better learning aspect term boundaries and improving model robustness;

\noindent
3) Removing the consistency training component results in a decrease of 1.57\% and 1.08\% in the average F1 score, respectively, indicating that narrowing the predicted probability distributions of aspect terms in different contexts can reduce the discrepancy of aspect terms with the same sentiment polarity in different semantic spaces.

\begin{table}[htbp]
\centering
\resizebox{\columnwidth}{!}{
\begin{tabular}{lccccc}
\toprule
& FR & ES & NL & RU & Avg \\
\midrule
\multicolumn{6}{c}{Based on mBERT:} \\
\midrule
w/o. Language Discriminator  & 49.70 & 60.61 & 51.57 & 52.21 & 53.52 \\
w/o. Consistency Training & 50.59 & 60.40 & 51.30 & 52.25 & 53.63 \\
\textbf{MSMO} & \textbf{51.42} & \textbf{63.26} & \textbf{52.68} & \textbf{53.45} & \textbf{55.20} \\
\midrule
\multicolumn{6}{c}{Based on XLM-R:} \\
\midrule
w/o. Language Discriminator & 59.82 & 68.10 & 62.41 & 60.99 & 62.83 \\
w/o. Consistency Training & 59.51 & 67.96 & 62.91 & 61.82 & 63.05 \\
\textbf{MSMO} & \textbf{61.01} & \textbf{69.47} & \textbf{63.26} & \textbf{62.52} & \textbf{64.13} \\
\bottomrule
\end{tabular}
}
\caption{The ablation study results.}
\label{tab:ablation_results}
\end{table}

\subsection{The Impact of the Parameter $\beta$}
\label{sec:setup_parameter}
\begin{figure}[ht]
\centering

\includegraphics[width=0.49\linewidth]{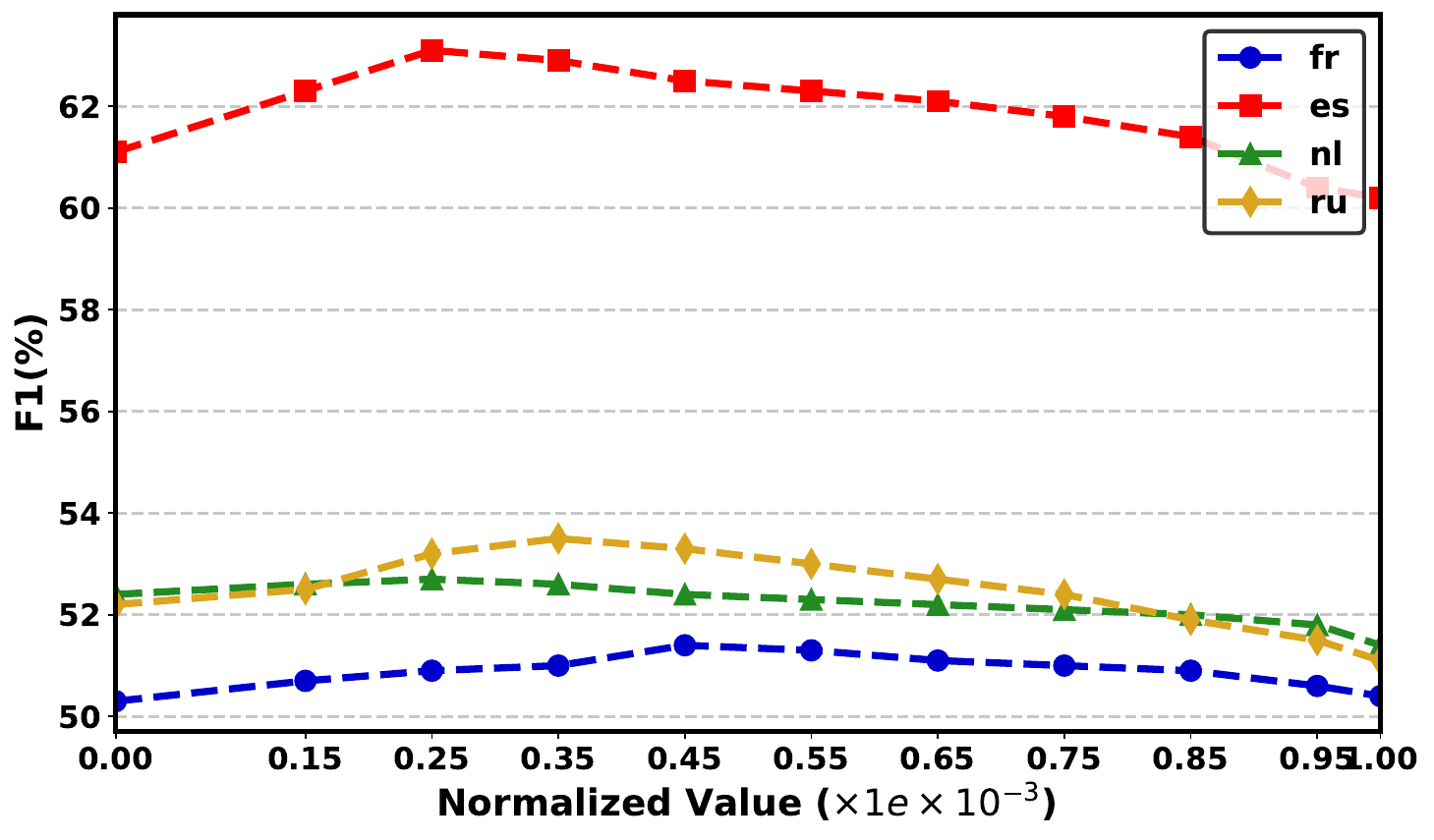}
\hfill
\includegraphics[width=0.49\linewidth]{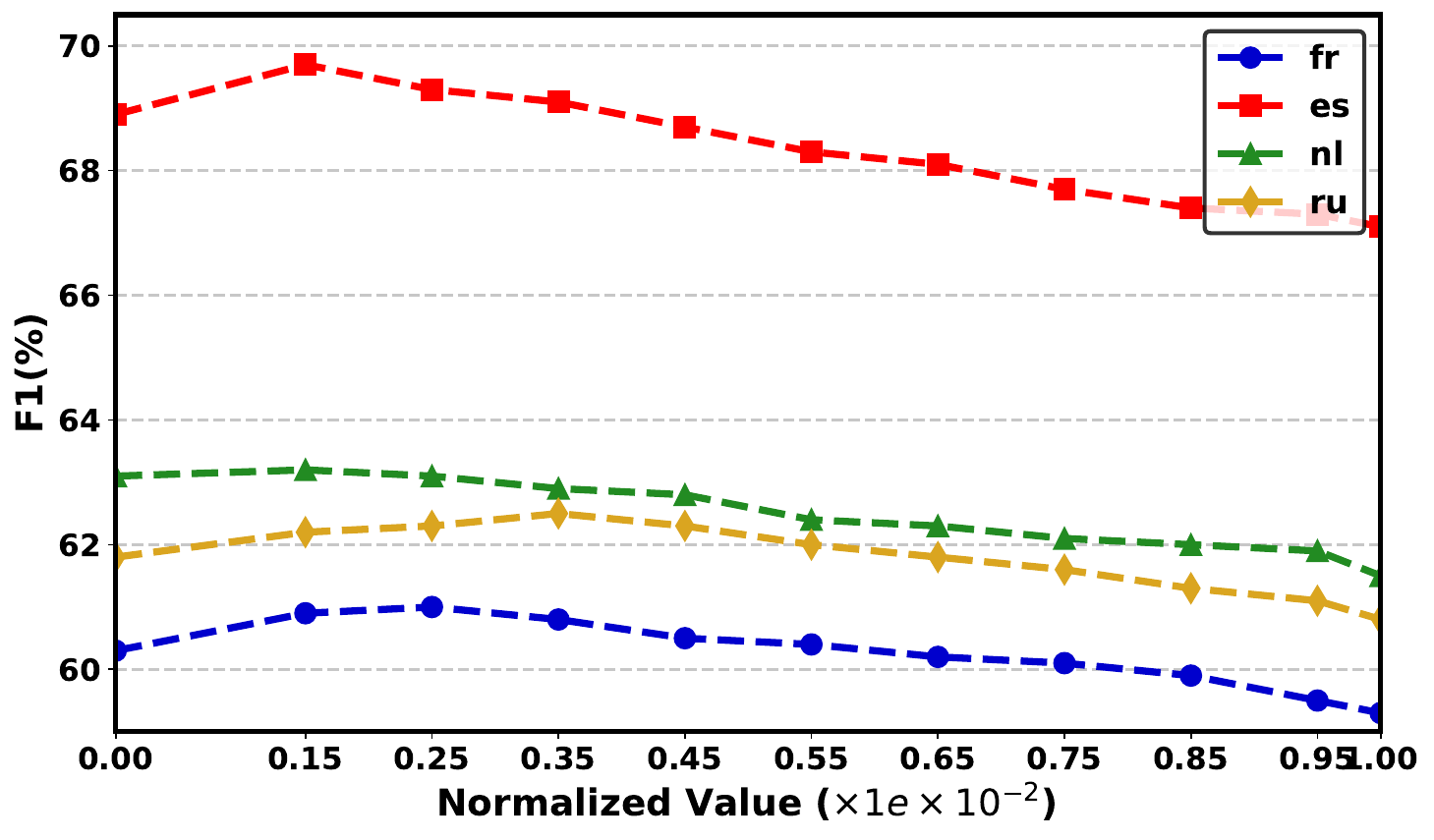}

\caption{The impact of the parameter $\beta$. Left: mBERT performance. Right: XLM-R performance.}
\label{fig:combined}

\end{figure}

In Figure \ref{fig:combined}, we present the impact of the parameter $\beta$ in Eq. \ref{eq:l} on the model performance.
We observe the following phenomena: 
1) Spanish achieves the best performance at lower values of $\beta$, which can be explained by the fact that similar languages share a closer semantic space and feature representations, making alignment easier. 
2) Both excessively large and small values of $\beta$ lead to performance degradation. When $\beta$ is too small, the model may overly rely on supervised training, resulting in poor cross-lingual generalization. On the other hand, when $\beta$ is too large, the model may focus too much on the consistency loss, neglecting language-specific details in sentiment analysis, which leads to suboptimal classification performance for certain languages. Therefore, the choice of $\beta$ should be adjusted according to the specific requirements of the task and the characteristics of the data, ensuring that the model can capture cross-lingual consistency without losing sensitivity to language-specific features.

\subsection{Excursion: Comparison with LLMs}
As recent advances in LLMs emerge, we also evaluate the LLM performance on the ABSA task to compare with our method. We evaluate the LLMs in two ways: by zero-shot prompting \citep{wu2024evaluating}, and by fine-tuning the LLMs on the source language (English) and then validating on target languages (similar to the supervised baseline). For zero-shot prompting, we apply the instruction-tuned version of GPT-4o \cite{gpt40modelcard}, Gemma-2 9B \cite{gemmateam2024gemma2improvingopen}, Llama-3.1 8B \cite{llama31modelcard}, and Mistral 7B \cite{jiang2023mistral7b}, and Qwen2.5 7B \cite{qwen2.5}. For fine-tuning, we employ the LoRA approach \cite{hu2022lora} on the four open-weight LLMs using the base models.
We present the results in Table \ref{tab:llm}. 
We apply the same prompt used in \citet{wu2024evaluating} to the models in zero-shot settings with only the prompt template as the system prompt. 

Still, our proposed MSMO method with multilingual distillation demonstrates a marked advancement in average performance, 
significantly outperforming the highest-scoring LLMs in zero-shot (GPT-4o) and in LoRA fine-tuning (Qwen-2.5). 
This suggests that while these LLMs excel in various NLP tasks, they may not be as effective for nuanced token-level classification tasks without additional fine-tuning 
\citep{wang2023gpt,nie2024decomposed}. 
Notably, performance across languages varies, with Spanish achieving relatively higher scores and French 
lower 
compared to other languages. This observation is consistent with the results obtained from our MSMO approach.

\begin{table}[htbp]
\centering
\scriptsize
\setlength\tabcolsep{3pt}
\renewcommand\arraystretch{0.85}
\begin{tabular}{lccccc}
\toprule
& FR & ES & NL & RU & Avg \\
\midrule
GPT-4o  & 48.43& 49.91& 49.94& 45.15& 48.36  \\
Gemma-2-9b-It   & 50.94& 48.80& 50.24& 39.34& 47.33  \\
Llama-3.1-8B-Instruct   & 23.15& 32.49& 33.53& 30.18& 29.84  \\
Mistral-7B-Instruct-v0.3 & 37.21& 38.32& 33.98& 26.58& 34.02 \\
Qwen2.5-7B-Instruct & 48.88 & 48.29 & 46.75 & 40.25 & 46.04 \\
\midrule
Gemma-2-9B + LoRA & 48.17 & 57.46 & 51.97 & 47.65 & 51.31 \\
Llama-3.1-8B + LoRA& 52.65 & 55.37 & 50.37 & 48.12 & 51.63 \\
Mistral-7B-v0.3 + LoRA & 59.46 & 60.79 & 56.89 & 50.52 & 56.92 \\
Qwen2.5-7B + LoRA & 63.01 & 68.95 & 60.84 & 53.50 & 61.58 \\
\midrule
\textbf{MTL-MSMO-DISTLL on mBERT} & \textbf{54.56} & \textbf{62.69} & \textbf{55.56} & \textbf{56.19} & \textbf{57.00} \\
\textbf{MTL-MSMO-DISTLL on XLM-R} & \textbf{63.23} & \textbf{70.95} & \textbf{66.24} & \textbf{64.36} & \textbf{66.20} \\
\bottomrule
\end{tabular}
\caption{Performance of different LLMs in zero-shot-prompting and in LoRA fine-tuning in comparison with MTL-MSMO-DISTLL (ours).}
\label{tab:llm}
\end{table}

\section{Related Work}
\textbf{Cross-Lingual ABSA.}
Research in cross-lingual ABSA (XABSA) generally falls into two categories: data alignment and embedding learning. Data alignment aims to incorporate language-specific knowledge into the target language, often using translation systems or dictionaries to convert annotated data from high-resource languages \cite{zhou-etal-2013-collective}. Techniques such as co-training \cite{zhou2015cl} and constrained SMT \cite{lambert-2015-aspect} improve data quality. Additionally, pre-trained multilingual word embeddings and methods like warm-up mechanisms \cite{li2021unsupervised} and shared vector spaces \cite{jebbara-cimiano-2019-zero} enhance model performance in multi-lingual ABSA, with additional CNN-based architectures offering further gains \cite{Wang2024}. \citet{zhang2021cross} adopt the translation-based methods by code-switching the aspect terms in target and source data for cross-lingual ABSA. Following this, \citet{lin2023} use contrastive learning for cross-lingual ABSA. There is also newly publicized multilingual ABSA datasets for implicit aspects \cite{wu2025mabsamultilingualdatasetaspectbased}.

\textbf{Adversarial Networks.} 
Adversarial training, popular in computer vision \cite{knoester2022}, has seen limited application in ABSA. Notable work includes \citet{Miyato2017VirtualAT}, who apply domain adversarial training to ABSA, and \citet{Wang_Pan_2018}, who use adversarial networks to align feature vectors across languages. Some methods have also explored character and word-level perturbations. \citet{Mamta2022AdversarialSG} generate adversarial samples for specific aspects while maintaining semantic coherence. Adversarial training helps evaluate model resilience and identify vulnerabilities \cite{lin2023}.

\textbf{Consistency Training.} 
Consistency training regularizes a model by ensuring predictions remain similar for both original and perturbed inputs \cite{zhou-etal-2022-conner}. While widely used in NLP, its application in ABSA is still emerging. Existing work includes \citet{chen-etal-2022-unsupervised-data}, which demonstrates that simple augmentations combined with consistency training yield competitive ABSA performance. Additionally, \citet{zhang-etal-2023-span} introduces a sentiment consistency regularizer to maintain sentiment consistency across spans.

\section{Conclusion}

In this work, we introduce the novel application of adversarial training and consistency training to cross-lingual aspect-based sentiment analysis. Our approach includes language discriminator and consistency training modules at the sentence and aspect levels, respectively, to better align aspect terms across languages. Multi-objective optimization further bridges semantic gaps between languages, establishing a robust baseline. Additionally, we demonstrate the effectiveness of knowledge distillation with the MSMO method. Extensive experiments confirm that our approach outperforms previous state-of-the-art methods. Future work will explore extending the MSMO framework to other multilingual NLP tasks.

\section*{Limitations}
Compared to the traditional cross-lingual ABSA methods, our proposed MSMO method incorporates different modules designed to learn the boundary features of aspect terms across different languages. However, the consistency of these features in highly diverse or idiomatic expressions may still present challenges, necessitating further refinement of these modules to handle more nuanced language variations. Additionally, our experiments rely on specific benchmark datasets, and whether our method can be generalized to other multilingual NLP tasks or real-world applications remains to be verified. Future work should include broader multilingual datasets to assess the robustness of our approach. We leave these for our future research.

\section*{Ethical Considerations}
This research was conducted in accordance with the ACM Code of Ethics. 
The dataset \cite{pontiki2016semeval} that we use is publicly available. 
We report only aggregated results in the main paper. 
We have not intended or do not intend to share any Personally Identifiable Data with this paper. 

\section*{Acknowledgments}
The authors acknowledge the use of ChatGPT exclusively to refine the text in the final manuscript and to assist in coding. ND and YH are supported in part by the National Science and Technology Major Project (2023ZD0121502).
CW and YX are supported in part by the Guangdong Basic and Applied Basic Research Foundation under Grant 2023A1515011370, the National Natural Science Foundation of China (32371114), the Characteristic Innovation Projects of Guangdong Colleges and Universities (No. 2018KTSCX049), and the Guangdong Provincial Key Laboratory (No. 2023B1212060076). XL was supported by Project for Improving Scientific Research Capacity of Key Construction Disciplines in Guangdong(No. 2025ZDJS022), Project of Teaching Quality and Teaching Reform in Undergraduate Universities of Guangdong Province: Modern Industrial College of Smart Agriculture in GPNU, Project of Guangdong Polytechnic Normal University (Grant No.22GPNUZDJS16).

\bibliography{custom}


\end{document}